\documentclass[runningheads]{llncs}

\usepackage{amsmath}
\usepackage{amssymb}
\usepackage{amsfonts}
\usepackage{pifont}
\usepackage{booktabs} 
\usepackage{hyperref}

\DeclareMathOperator*{\argmin}{argmin}
\DeclareMathOperator*{\argmax}{argmax}
\DeclareMathOperator*{\aggregate}{aggregate}
\DeclareMathOperator*{\combine}{combine}
\DeclareMathOperator*{\activate}{activate}
\DeclareMathOperator{\st}{s.t.}

\usepackage{graphicx}
\usepackage{xcolor}
\usepackage{multirow}
\newcommand{\matheus}[1]{{ #1}}

\begin{document}
\title{Neural Architecture Search in \\Graph Neural Networks}
%
%
\author{Matheus Nunes \and Gisele L. Pappa}
%
\authorrunning{Nunes, M and Pappa, G. L.}
%
\institute{Universidade Federal de Minas Gerais, Minas Gerais, Brazil
\email{\{mhnnunes,glpappa\}@dcc.ufmg.br}
}
\maketitle              
\begin{abstract}
 Performing analytical tasks over graph data has become increasingly interesting due to the ubiquity and large availability of relational information. However, unlike images or sentences, there is no notion of sequence in networks. Nodes (and edges) follow no absolute order, and it is hard for traditional machine learning (ML) algorithms to recognize a pattern and generalize their predictions on this type of data. Graph Neural Networks (GNN) successfully tackled this problem. They became popular after the generalization of the convolution concept to the graph domain. However, they possess a large number of hyper-parameters and their design and optimization is currently hand-made, based on heuristics or empirical intuition. Neural Architecture Search (NAS) methods appear as an interesting solution to this problem. In this direction, this paper compares two NAS methods for optimizing GNN: one based on reinforcement learning and a second based on evolutionary algorithms. Results consider 7 datasets over two search spaces and show that both methods obtain similar accuracies to a random search, raising the question of how many of the search space dimensions are actually relevant to the problem. 

\end{abstract}
%
%
%

\section{Introduction} \label{sec:introduction}

Performing analytical tasks over graph\footnote{In this work we use the terms ``graph'' and ``network'' interchangeably. When referring to ``neural networks'' we will use NN or ``neural network''.} data has become increasingly interesting due to the ubiquity and large availability of relational information. Predicting interaction between proteins,
classifying users in social networks
and recommending movies to users
are some classical examples of such tasks \cite{zhang2020gnn}. However, unlike images (formed by a grid of pixels) and sentences (formed by a string of ordered words), there is no notion of sequence in networks. Nodes (and edges) follow no absolute order, so it is hard for traditional machine learning (ML) algorithms, which were built to handle data stored in tensors, to recognize a pattern and generalize their predictions on this type of data \cite{zhang2018end}.
%

Due to the success of convolutional neural networks (CNNs) for tasks such as image classification \cite{he2016deep}, object identification \cite{liu2016ssd} and semantic segmentation \cite{badrinarayanan2017segnet}, a large body of work began to re-define the concept of convolution to the graph domain. Following the work of Gori et. al. \cite{gori2005graphlearning} and Scarselli et al. \cite{scarselli2009neural} on Graph Neural Networks (GNNs), the concept of spectral-based graph convolution function was defined by Bruna et al. \cite{bruna2014spectral} and later refined by Defferrard et. al. \cite{deferrard2016localized}.
In this approach, unlike traditional neural networks where the architecture is composed by fully connected layers of neurons, graph neural networks follow the graph structure itself \cite{scarselli2009neural}. Forward propagation is done on the nodes of the graph, which pass information onto the next layer by aggregating information from the neighborhood and applying an activation function to the result. 

Since the concept of convolution was adapted to the context of graphs, a plethora of GNN models were proposed, including GraphSAGE \cite{hamilton2017inductive}, Graph Attention Networks (GAT) \cite{velickovic2018gat}, Graph Isomorphism Network (GIN) \cite{xu2018powerful} and many others. These methods achieve state-of-the-art results on tasks such as node classification and link prediction. However,
the design and optimization of GNN architectures is currently hand-made, based on heuristics or empirical intuition, which makes it an ineffective and error prone task \cite{xu2018powerful}.

Automated Machine Learning (AutoML) appears as a solution to this problem, as it aims to automate the process of building and optimizing machine learning pipelines, relieving users from that burden \cite{elshawi2019automated}.
Neural Architecture Search (NAS) is considered the 
current challenge in automating machine learning algorithms \cite{elsken2019neural}.
Its methods are composed by a search space of possible architectures, a search method to explore this space and an evaluation framework for the generated architectures.


To the best of our knowledge, there were few attempts in the literature to employ NAS for GNNs \cite{gao2019graphnas,zhou2019autognn}. In these works, reinforcement learning methods are used to explore similar search spaces. The NAS literature poses two main types of methods as the most effective to solve the problem: reinforcement learning (RL) and evolutionary algorithms (EAs) \cite{elsken2019neural}. The second type of technique has been so far overlooked in the context of GNNs.

This work employs an EA previously proposed for NAS in the context of image classification \cite{real2019aging} to optimize GNNs and performs a comparative analysis of the method with reinforcement learning and random search in terms of model accuracy and runtime. It also conducts a study of the characteristics of the previously proposed search spaces for GNNs in order to identify opportunities for performance improvement on GNN NAS algorithms. Results show that {both RL and EA are able to find equivalent models in terms of accuracy}, with {EA being faster} in some cases, which corroborates previous findings for image classification. Furthermore, following the already discussed problems of large search spaces -- such as those required in the case of GNNs --  with many low effective dimensions \cite{bergstra2012random}, we show a {Random Search is able to find architectures with equivalent accuracy while being faster}.
We discuss these results in the light of previous works that discuss this problem.

The remainder of this work is organized as follows. Section 2 introduces background on GNNs and Section 3 discusses related work. Section 4 describes the methodology followed to apply the tested methods in GNN search spaces, while Section 5 presents the results. Finally, Section 6 draws conclusions and discusses directions of future work.



\section{Background} \label{sec:background}

In this work, we assume as input a graph composed of a set of nodes and edges, $G = (N,E)$. Each node $n_i \in N$ is attached to a feature/attribute vector $x_i \in X$, and a label $l_i \in L$. The presence of \matheus{node} labels indicates that we are assuming a \textbf{supervised learning} situation. 
We define by $\mathcal{N}(i)$ the neighborhood of a node $i$, i.e., the set of nodes connected to $i$ by an edge. 
The primary concept behind GNNs is that each node in the graph represents an abstract concept, and edges represent the relationship between these concepts. Therefore, the node's features should correlate with its neighboring features, defining a state (or hidden node representation) $h_i \in \mathcal{H}_N$ for each node \cite{scarselli2009neural}.

Traditionally, each GNN layer is composed of a function that aggregates information from the neighborhood of each node $\mathcal{N}(i)$, forming an intermediate vector $h_{\mathcal{N}(i)}$, and a second function that combines this value with the current node representation $h_{i}$, which in turn goes through an activation function before being output \cite{scarselli2009neural,hamilton2017inductive}. Formally, this process can be defined as:

\begin{align} \label{eq:gnn_formulation}
h_{\mathcal{N}(i)}^{(k)} = \aggregate({h_j^{k-1} : j \in \mathcal{N}(i)}) \\
h_{i}^{(k)} = \activate(\combine(h_{i}^{(k-1)}, h_{\mathcal{N}(i)}^{(k)}))
\end{align}

\begin{figure}[th]
\includegraphics[width=\textwidth]{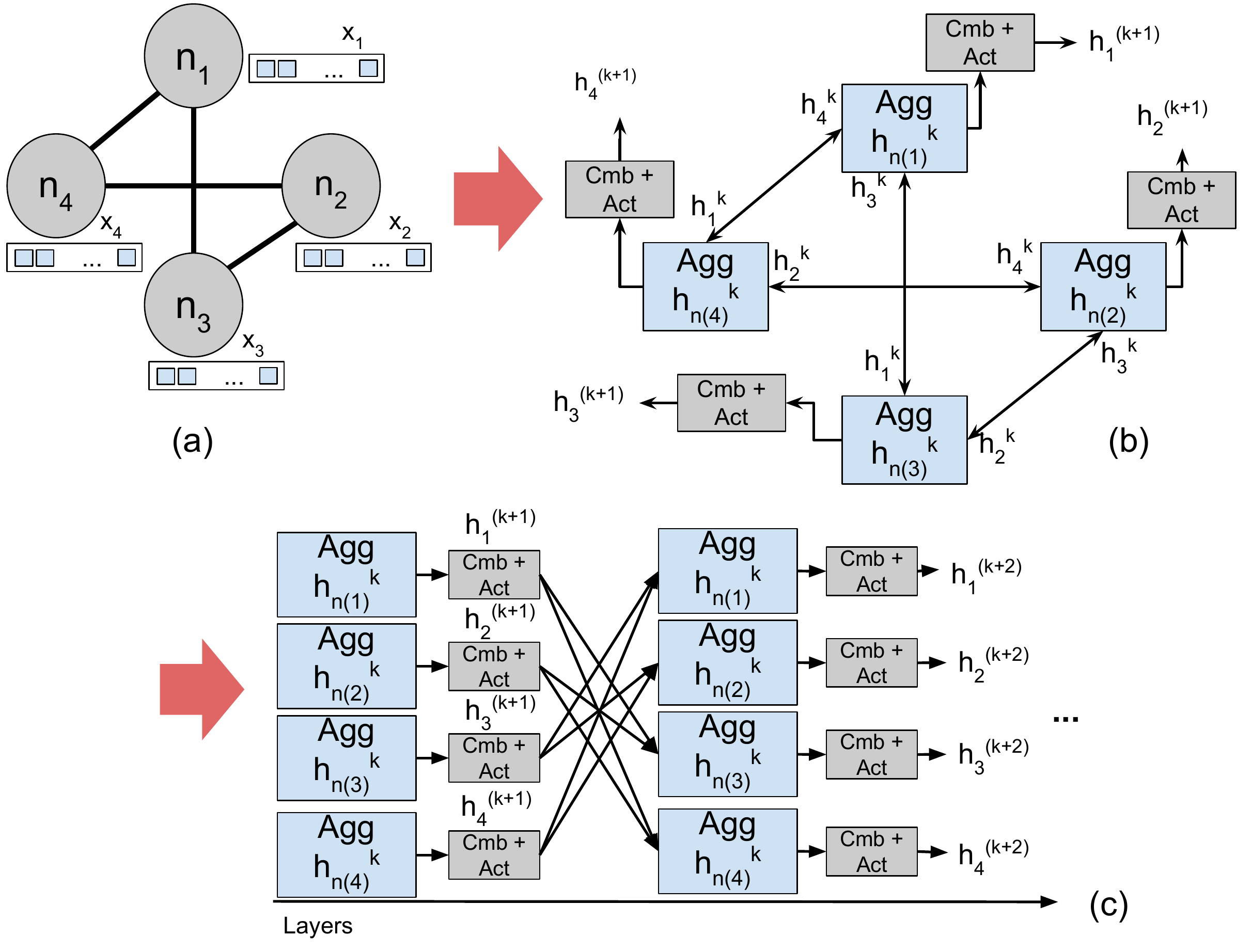}
\caption{Structure of a GNN, adapted from Scarselli et. al. \cite{scarselli2009neural}}
\label{fig:GNN_struct}
\end{figure}

By convention, the first hidden representation of each node is its feature vector, $h_i^{(0)} = x_i$ \cite{kipf2017gcn}. Figure \ref{fig:GNN_struct} shows how the structure of a GNN is generated. Given the graph represented in part (a) of the figure, which has 4 nodes $n_i$ and a feature vector $x_i$ associated to each of them, an intermediate representation is generated ((b) in the figure). In this representation, for each node, the neighborhood information generates the intermediate vectors $h_i$ according to the process described in Eq.
~\ref{eq:gnn_formulation}. The third part of the picture (c) shows the GNN itself, where each layer corresponds to an update of the state of the feature vectors of the current node.

In this work we consider undirected graphs and a one-hop neighborhood for each node, which means that only features from a node's direct neighbors are considered in aggregation. There are many options of aggregation and activation functions, and other mechanisms can also be added to this standard GNN architecture. These components choices are the main subject of this paper, as detailed in Section~\ref{sec:search_spaces}.

\section{Related Work} \label{sec:related_work}

NAS is considered the 
current challenge
in automating machine learning algorithms, after the success of automated feature engineering \cite{elsken2019neural}. Famous NAS works can be roughly split into two categories: Reinforcement Learning (RL) \cite{zoph2018image,cai2018nas}  and Evolutionary Algorithms (EA) \cite{real2019aging}. It has been shown that both types of methods are able to find models that perform better than hand-crafted engineered ones, but Real et al. presents empirical proof that EA-based and RL-based methods are able to find equally well-suited models in terms of performance, with EA-based methods finding less complex models in less overall time \cite{elsken2019neural,real2019aging}. 
Our idea is to adapt and employ NAS methods to the task of finding a good GNN model for large-scale graph embedding, whereas in previous works, the tasks of interest were mostly image classification and object detection.



To the best of our knowledge, NAS has not yet been \matheus{largely} explored in the context of GNNs. GraphNAS \cite{gao2019graphnas} is one of the few that uses RL to find feasible architectures for the node classification task. The authors define a search space composed of sampling, aggregation and gated functions, which can be extended to account for hyperparameters.  
Auto-GNN \cite{zhou2019autognn} follows the same line of work, exploring RL and  a similar search space to GraphNAS.

\begin{figure}[ht]
    \centering
    \includegraphics[width=0.9\textwidth]{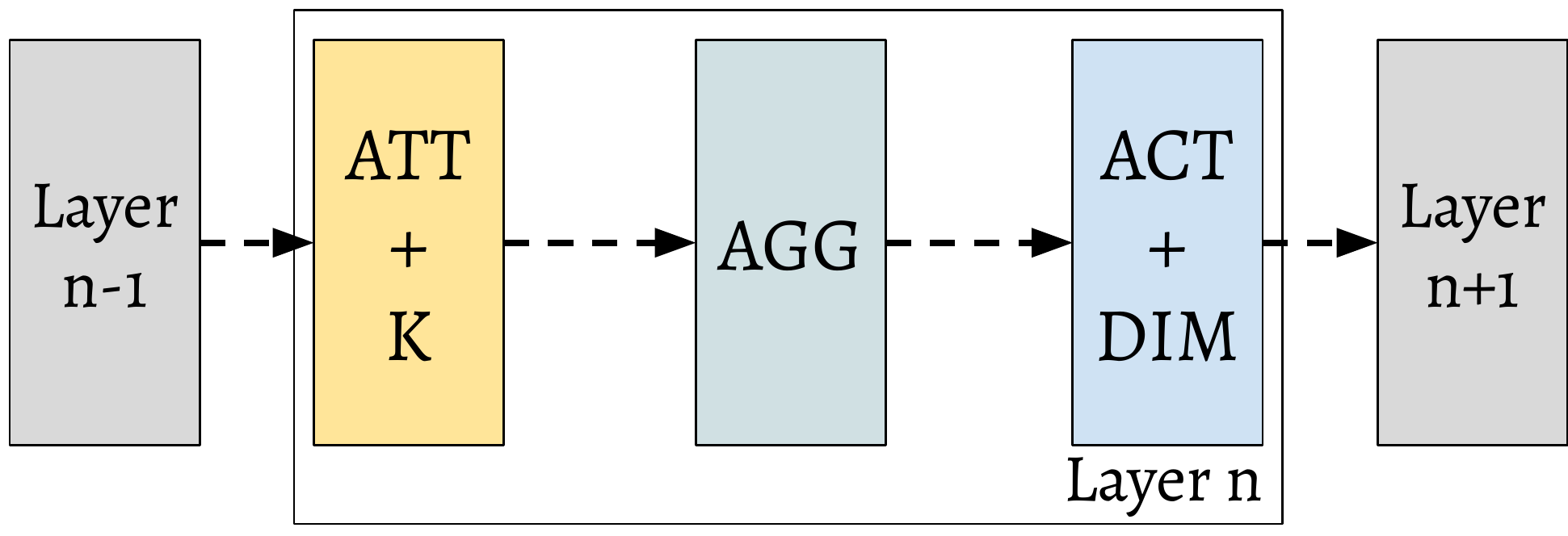}
    \caption{Macro Search Space GNN Layer Example}
    \label{fig:gnn_layer}
\end{figure}

\section{Methodology} \label{sec:methodology}



The problem of NAS in GNNs can be formally defined as follows.
Given a dataset $\mathcal{D}$ -- split into training and validation sets $\mathcal{D}_{train}$ and $\mathcal{D}_{valid}$, respectively -- and a search space of Graph Neural Architectures $\mathcal{A}$, capable of generating a GNN with an architecture $a \in \mathcal{A}$ with its own set of hyperparameters $\Lambda$, 
the goal is to find the model with the highest expected accuracy $\mathcal{E}$ on $\mathcal{D}_{valid}$, when its parameters $w^*$ are set on $\mathcal{D}_{train}$, setting the following bi-level optimization problem:
\[
\begin{aligned} \label{eq:gnn_nas_formulation}
\argmax_{a_{\lambda} \in \mathcal{A}, \lambda \in \Lambda, w^*} & \mathcal{E}[\mathcal{}(a_{\lambda}(w^*, \mathcal{D}_{valid}))] \\
    & \st w^* =  \argmin_{w}  \mathcal{L}(a_\lambda(w, \mathcal{D}_{train})),
\end{aligned}
\]

This section details the search spaces $\mathcal{A}$ previously defined for GraphNAS \cite{gao2019graphnas} and describes the evolutionary algorithm and the RL methods we evaluated in the context of GNN architecture search.


\begin{table}[t]
\centering
\caption{Macro search space options for 5 actions.}
\label{tab:macro_space_actions}
\begin{tabular}{llr}
\hline
{$\mathbf{ATT}$}          & {$\mathbf{AGG}$} & {$\mathbf{ACT}$}    \\ 
\hline
const,  $e_{ij}=1$       &  sum     & tanh     \\
gcn,    $e_{ij}=1/d_id_j$      & mean   & linear  \\
gat,  $e_{ij}=leaky\_relu((W_lh_i + W_rh_j))$      & max    & softplus       \\
sym-gat,  $e_{ij}=e_{ji} + e_{ij}$     & mlp    &    sigmoid     \\
cos,  $e_{ij}=<W_lh_i, W_rh_j>$         &       &  elu        \\
linear,  $e_{ij}=tanh(sum(W_lh_j))$       &     &    relu    \\
gen\_linear,  $e_{ij}=W_atanh(W_lh_i + W_rh_j)$ &       & relu6 \\
      &          & leaky\_relu \\\hline     
$\mathbf{K}$ &\multicolumn{2}{l}{$2^i$, $i \in \{1,...,6\}$}\\
$\mathbf{DIM}$ & \multicolumn{2}{l}{$2^i$, $i \in \{2,...,8\}$}\\\hline
\end{tabular}
\end{table}

\subsection{Search Spaces} \label{sec:search_spaces}

The two search spaces evaluated in this work, named by the authors in \cite{gao2019graphnas} as ``Macro'' and ``Micro'', are composed by different GNN layers, as detailed next.  

\subsubsection{Macro Search Space} \label{sec:macro_search_space}

The name ``Macro'' comes from the fact that architectures generated from this space always follow the same structure: each layer is composed by a multi-head attention mechanism ${ATT}$ and the number of heads ${K}$, a choice of aggregator ${AGG}$, the output dimension ${DIM}$ and an activation function ${ACT}$, in this order. The neighborhood sampling method is fixed as a first-order sampler, i.e. only direct neighbors of each node are sampled at each step. 

Considering the definitions in Section~\ref{sec:background}, we have a new component here, which is the attention mechanism. As described by the authors in \cite{velickovic2018gat}, an attention mechanism -- implemented by the coefficients $e_{ij}$, is designed to attribute different importance value to the features of each of a node's neighbors. \matheus{Such} coefficients are calculated only for $j \in \mathcal{N}(i)$ for performance reasons (in order to avoid an $N \times N$ matrix), \matheus{and} in practice define the importance of node $j$'s features over node $i$. They are implemented as a single-layer feed-forward neural network, and a range of options to this mechanism is available (see first column of Table~\ref{tab:macro_space_actions}). Multi-head attention is a way of having independent attention mechanisms over the node's features. It has been proven that concatenating the results of these independent mechanisms yields better results than using a single attention head \cite{velickovic2018gat}. 

Figure \ref{fig:gnn_layer} presents the disposition of the actions. The number of multi-heads ${K}$ can be merged with the attention mechanism ${ATT}$ as they alter the same behavior. The output dimension ${DIM}$ can also be merged with the activation function ${ACT}$.

Table \ref{tab:macro_space_actions} presents the options for each action on the layers. Considering the number of options for each action on the layers, the search space presents ($7 \times 6 \times 4 \times 7 \times 8$) = $9408$ possibilities for each layer. According to the authors in \cite{kipf2017gcn}, GNNs achieve the best overall results using architectures with 2 or 3 layers. Therefore in this paper the architectures have 2 layers, in a total of $9408^2 = 88,510,464$ architecture possibilities.


One important characteristic of this search space is that the hyperparameters of the GNNs, such as learning rate, dropout, weight decay are kept fixed. The learning rate is set to $0.005$, the dropout to 0.6 and the weight decay to $5\times10^{-4}$.

\begin{table}[t]
    \centering
        \caption{Micro search space action and hyperparameters.}
    \label{tab:micro}

    \begin{tabular}{ll}\hline
         $\mathbf{CNV}$& $GAT_{1,...,8}$, GCN, Cheb, SAGE, ARMA, SG, Linear, Zero   \\
         $\mathbf{CMB}$&  Add, Product, Concat \\
         $\mathbf{ACT}$&  Sigmoid, tanh, elu, relu, linear  \\
         \textbf{LR}& $\{ 1\times10^{-2}, 1\times10^{-3},  1\times10^{-4}\}$\\
         \textbf{DO}& $\{0.0, 0.1, ..., 0.9\}$\\
         \textbf{WD}& $\{0, 1\times10^{-3}, 1\times10^{-4}, 5\times10^{-4}, 1\times10^{-5}, 5\times10^{-5}\}$\\
         \textbf{HU}& $2^i, i \in \{3,...,9\}$\\\hline
    \end{tabular}
\end{table}

\subsubsection{Micro Search Space} \label{sec:micro_search_space}
The name ``Micro'' comes from the fact that architectures generated from this search space are composed by combining different convolution schemes, and do not follow a single fixed structure. The choice of actions in this space are: a convolutional layer ${CNV}$, a combination scheme ${CMB}$ and an activation function ${ACT}$. The hyperparameters which can be tuned are: the learning rate ${LR}$, the dropout rate ${DO}$, the weight decay rate ${WD}$ and the number of hidden units ${HU}$. In the options for ${CNV}$, the option $GAT_{1,...,8}$ means that there are 8 possible $GAT$ convolutions, using 1 to 8 multi-heads attention.


\begin{figure}
    \centering
    \includegraphics[width=0.9\textwidth]{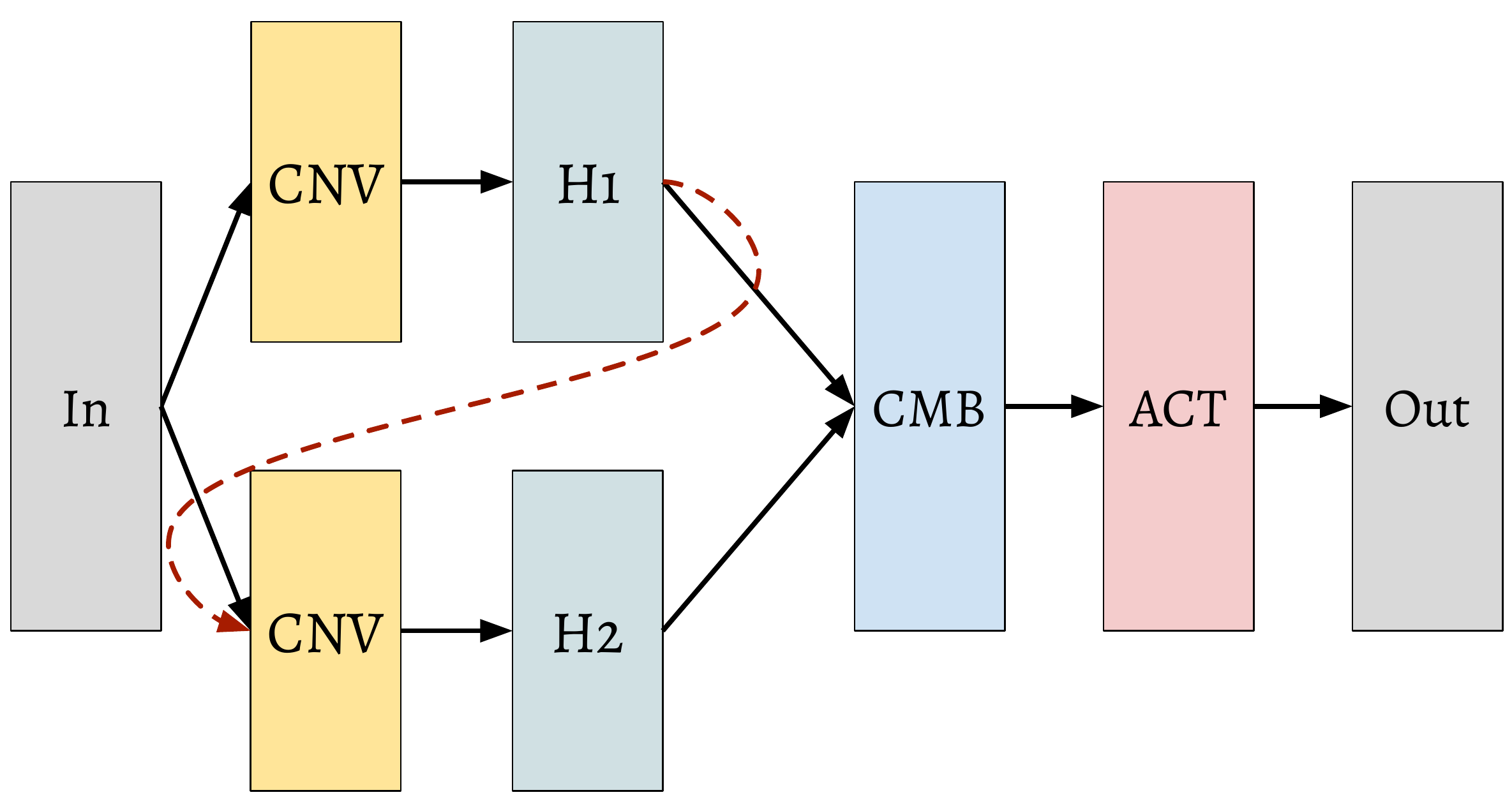}
    \caption{Micro Search Space GNN architectures Example}
    \label{fig:gnn_micro}
\end{figure}

Figure \ref{fig:gnn_micro} illustrates the types of architectures that can be generated from this space. The straight arrows represent one type of connectivity, where the input is fed to two separate convolutional layers and their outputs are fed to the combination layer. The dashed line represents the second type, when two convolutional layers are stacked before feeding the output to the combination layer. The full list of actions and hyperparameters for this space is presented in Table
~\ref{tab:micro}. \matheus{Regarding the number of possibilities for each action and hyperparameter listed, there are $(15 \times 15 \times 3 \times 5 \times 3 \times 10 \times 5 \times 7) = 3,543,750$ architecture possibilities in this space.}

Note that the architectures in the micro-space take advantage of convolutions.
Graph convolution methods are classified mainly into two streams, both covered by the micro-search space: {spectral-based} and {spatial-based} methods \cite{wu2019comprehensive}. Spectral methods \cite{bruna2014spectral,kipf2017gcn} rely on spectral properties of the graph, by finding eigenvectors of the normalized graph Laplacian. This approach is limited because eigendecomposition is an expensive operation, eigenbasis are sensible to minimal graph perturbations and the learned filters do not generalize well to graphs of different structure (therefore they do not work well on inductive learning scenarios). Spatial-based methods \cite{hamilton2017inductive,velickovic2018gat} follow the message passing idea of traditional GNNs (also known as Recursive GNNs), in which a node's hidden representation is an input to its neighbors computation. These methods are scalable to large graphs and are more generalizable to various types of graphs (heterogeneous, directed, graphs which contain edge labels, etc.).



\subsection{Search Methods} \label{sec:search_methods}

This section describes the two methods we apply to search the macro and micro search spaces described in the previous section: the evolutionary method and the reinforcement learning. We also describe the random search method that will be used as a baseline for the results.

\vspace{0.2cm}
\noindent \textbf{Evolutionary algorithm} -
    Evolutionary methods are inspired by Darwin's theory of evolution, and evolve a set of individuals -- which represent solutions to the problem at hand -- for a number of iterations (also known as generations) \cite{eiben2015}. From one iteration to the next, individuals are evaluated according to a fitness function, which assesses their ability to solve the problem. The value of fitness is used to probabilistic select the individuals that will undergo crossover and mutation operators, which are applied according to user-defined probabilities.
    We explore an evolutionary method inspired on the Aging Evolution method, described by Real at. al. \cite{real2019aging}. In this method, a population of individuals --i.e., a set of GNNs -- is generated randomly by sampling options for each action in a layer, considering the number of layers specified. 
    These GNNs are then trained in a training set and have their accuracy measured on a validation set. This value of accuracy is used to select an individual via tournament selection to generate a new offspring. The child individual is generated via mutation, which is uniform over the actions and replaces the selected action by a random option. The child individual is always added to the population and the oldest individual in the population (i.e., the individual that has been in the population for the highest number of iterations) is {always} removed (hence the name 
   ``Aging Evolution"). 


\vspace{0.2cm}
\noindent\textbf{Reinforcement Learning} - GraphNAS uses a LSTM (Long-Short Term Memory) network as a controller to generate fixed-length architectures, which act as GNN architecture descriptors and can be viewed as a list of actions. The accuracy achieved by the GNN in the validation dataset at convergence is used as the reward signal to the training process of the reinforcement learning controller.
As the reward signal $\mathcal{R}$ is non-differentiable, a policy gradient method is used to iteratively update $\theta$ with a moving average baseline for reward to reduce variance. 

\vspace{0.2cm}
\noindent \textbf{Random Search} - An initial random GNN is generated by sampling options from each action in a layer, for the specified number of layers. The GNN is trained and the accuracy on the validation set measured. This process is repeated for the specified number of iterations, \matheus{storing the GNN with the highest accuracy.}

\section{Experimental Analysis} \label{sec:experiments}

We \matheus{assess} the performance of the evolutionary algorithm (EA)\footnote{Code available at: \url{https://github.com/mhnnunes/nas_gnn}}, the reinforcement learning (RL) method and the random search (RS) on the transductive learning scenario, \matheus{in a node classification task}, over a set of 7 datasets in terms of accuracy and runtime, as detailed next.
\matheus{It is important to note that this work does not compare the architectures obtained by the optimization methods to hand-crafted ones, as that was already done in GraphNAS' paper \cite{gao2019graphnas}.}

\subsection{Datasets} \label{sec:datasets}

Table \ref{tab:datasets} presents the details of the datasets, as
previously used in \cite{shchur2018pitfalls} and
provided by \texttt{Pytorch Geometric}\footnote{\url{https://github.com/rusty1s/pytorch_geometric}}. 
For all cases, we are dealing with a \matheus{node} classification task, where we use information from the nodes with known-labels to assign a class to nodes with unknown label (test set).

\setlength{\tabcolsep}{3.5pt}
\begin{table}[ht]
\centering
\caption{Dataset characteristics.}
\label{tab:datasets}
\begin{tabular}{lrrrr}\hline
\textbf{Dataset (Abbrv.)}                & \textbf{\# Classes} & \textbf{\# Features} & \textbf{\# Nodes} & \textbf{\# Edges} \\ \hline
\textbf{CORA} (\textbf{COR})             & $7$                   & $1433$                 & $2708$              & $10556$             \\
\textbf{Citeseer} (\textbf{CIT})         & $6$                   & $3703$                 & $3327$              & $9104$              \\
\textbf{Pubmed} (\textbf{MED})           & $3$                   & $500$                  & $19717$             & $88648$             \\
\textbf{Coauthor CS (CS)}                & $15$                  & $6805$                 & $18333$             & $163788$            \\
\textbf{Coauthor Physics} (\textbf{PHY}) & $5$                   & $8415$                 & $34493$             & $495924$            \\
\textbf{Amazon Computers} (\textbf{CMP}) & $10$                  & $767$                  & $13752$             & $491722$            \\
\textbf{Amazon Photo}  (\textbf{PHO})    & $8$                   & $745$                  & $7650$              & $238162$            \\ \hline
\end{tabular}
\end{table}

The first three datasets (\textbf{COR}, \textbf{CIT}, \textbf{MED}) are paper co-authorships networks, used previously in \cite{kipf2017gcn}. Nodes represent documents, and an edge between two documents means that one paper cited the other. Class labels represent sub-areas of machine learning \cite{sen2008collective}. Node features are sparse bag-of-words vectors. 

\textbf{CS} and \textbf{PHY} are also co-authorship networks, based on the Microsoft Academic Graph from KDD Cup 2016. However, in these datasets nodes represent authors instead of papers, connected by an edge if they have co-authored a paper. Node features represent paper keywords for each author's papers. Class labels indicate the most active field of study for each author in the network. 

\textbf{CMP} and \textbf{PHO} are segments of the Amazon co-purchase graph, where nodes represent products and edges are added between items frequently bought together. The nodes features are a bag-of-words representation of product reviews, and class labels represent the product category.

\begin{table}[ht]
    \centering
    \caption{Accuracies and execution times (in $\times 10^4$ seconds) of search methods.}
    \label{tab:acc_time_table}
    \begin{tabular}{llcc|cc} \hline
      	         & &\multicolumn{2}{c}{\textbf{Macro}}     & \multicolumn{2}{c}{\textbf{Micro}}  \\ \hline
      	         & & \textbf{Accuracy} & \textbf{Time} & \textbf{Accuracy} & \textbf{Time} \\ \hline
                & \textbf{EA} & $0.83 \pm 0.007$  & $0.75 \pm 0.16$ &  $0.82 \pm 0.005$  &  $1.73 \pm 0.53$  \\
\textbf{COR}    & \textbf{RL} & $0.83 \pm 0.003$  & $1.45 \pm 0.38$ &  $0.81 \pm 0.001$  &  $2.42 \pm 0.62$  \\
                & \textbf{RS} & $0.82 \pm 0.003$  & $0.96 \pm 0.02$ &  $0.80 \pm 0.009$  &  $1.20 \pm 0.21$  \\  \hline
                & \textbf{EA} & $0.75 \pm 0.002$  & $1.18 \pm 0.10$ &  $0.71 \pm 0.007$  &  $2.80 \pm 0.72$  \\
\textbf{CIT}    & \textbf{RL} & $0.73 \pm 0.004$  & $1.52 \pm 0.42$ &  $0.68 \pm 0.006$  &  $2.24 \pm 0.08$  \\
                & \textbf{RS} & $0.73 \pm 0.005$  & $1.05 \pm 0.03$ &  $0.69 \pm 0.006$  &  $1.29 \pm 0.04$  \\  \hline
                & \textbf{EA} & $0.82 \pm 0.003$  & $1.40 \pm 0.37$ &  $0.82 \pm 0.009$  &  $1.40 \pm 0.09$  \\
\textbf{MED}    & \textbf{RL} & $0.80 \pm 0.003$  & $2.10 \pm 0.14$ &  $0.76 \pm 0.017$  &  $2.58 \pm 0.28$  \\
                & \textbf{RS} & $0.85 \pm 0.045$  & $1.31 \pm 0.02$ &  $0.80 \pm 0.009$  &  $1.10 \pm 0.18$  \\  \hline
                & \textbf{EA} & $0.98 \pm 0.001$  & $3.35 \pm 0.78$ &  $0.99 \pm 0.002$  &  $2.65 \pm 0.48$  \\
\textbf{CS}     & \textbf{RL} & $0.95 \pm 0.001$  & $3.13 \pm 0.11$ &  $0.97 \pm 0.002$  &  $2.90 \pm 0.34$  \\
                & \textbf{RS} & $0.97 \pm 0.001$  & $1.50 \pm 0.03$ &  $0.99 \pm 0.001$  &  $1.58 \pm 0.05$  \\  \hline
                & \textbf{EA} & $0.99 \pm 0.002$  & $4.21 \pm 0.85$ &  $0.99 \pm 0.000$  &  $1.53 \pm 0.15$  \\
\textbf{PHY}    & \textbf{RL} & $0.98 \pm 0.001$  & $3.34 \pm 0.27$ &  $0.98 \pm 0.001$  &  $2.01 \pm 0.19$  \\
                & \textbf{RS} & $0.98 \pm 0.001$  & $2.08 \pm 0.07$ &  $0.99 \pm 0.001$  &  $1.11 \pm 0.05$  \\  \hline
                & \textbf{EA} & $0.91 \pm 0.005$  & $3.09 \pm 0.49$ &  $0.93 \pm 0.004$  &  $4.02 \pm 1.94$  \\
\textbf{CMP}    & \textbf{RL} & $0.90 \pm 0.010$  & $3.43 \pm 0.21$ &  $0.92 \pm 0.008$  &  $3.68 \pm 0.27$  \\
                & \textbf{RS} & $0.89 \pm 0.004$  & $1.69 \pm 0.07$ &  $0.92 \pm 0.002$  &  $2.05 \pm 0.07$  \\  \hline
                & \textbf{EA} & $0.97 \pm 0.002$  & $2.48 \pm 0.22$ &  $0.98 \pm 0.004$  &  $1.66 \pm 0.41$  \\
\textbf{PHO}    & \textbf{RL} & $0.96 \pm 0.005$  & $3.65 \pm 0.19$ &  $0.97 \pm 0.002$  &  $1.88 \pm 0.23$  \\
                & \textbf{RS} & $0.96 \pm 0.002$  & $1.82 \pm 0.04$ &  $0.97 \pm 0.002$  &  $1.08 \pm 0.04$  \\ \hline
    \end{tabular}
\end{table}

\subsection{Experimental Setup}

All search methods were executed for 1000 iterations in order to enable a fair comparison. \matheus{In each iteration, a single GNN architecture is generated, trained on $\mathcal{D}_{train}$ and evaluated (in terms of accuracy) on $\mathcal{D}_{valid}$. The architecture with the highest validation accuracy is saved across iterations, and returned as the result of the optimization process. The generated architectures are trained using the following fixed hyperparameters for all search spaces and methods: minimizing cross-entropy loss using ADAM optimizer, initial learning rate of 0.005 and an early stopping strategy with a patience of $100$ epochs.}

Random search has only one parameter: the number of iterations.
The reinforcement learning controller is trained using the same hyperparameters as described on GraphNAS' paper \cite{gao2019graphnas}: a one-layer LSTM with 100 hidden units, ADAM optimizer, learning rate at $3.5\times 10^{-4}$ and random initialization of weights.
Aging Evolution has three main parameters: the population size, the tournament size $k$ and the number of iterations $n$. 
The first parameter is related to the number of solutions evaluated during the search process, while the tournament size controls the convergence speed. The higher the value of $k$, the faster the algorithm converges. From all tested values ($\{100,25\}, \{25,2\}, \{100,3\}$), the best results were achieved using 
the population size set to $100$ and $k$ set to 3. 

The \matheus{dataset split} between training, validation and testing sets was done in the same way as in the GraphNAS public code\footnote{\url{https://github.com/GraphNAS/GraphNAS}}: the last $1000$ nodes are separated for validation and testing, split evenly between the two.

All experiments were repeated 5 times as the methods are non-deterministic. The experiments were run on a machine with a 16-core Intel(R) Xeon(R) Silver 4108 CPU @ 1.80GHz, 16GB DIMM DDR4 @ 2666 MHz RAM, and a NVIDIA GV100 [TITAN V] graphics card, with 12GB dedicated RAM.

\subsection{Results} \label{sec:exp_setup}

\begin{figure}[ht]
    \includegraphics[width=\textwidth]{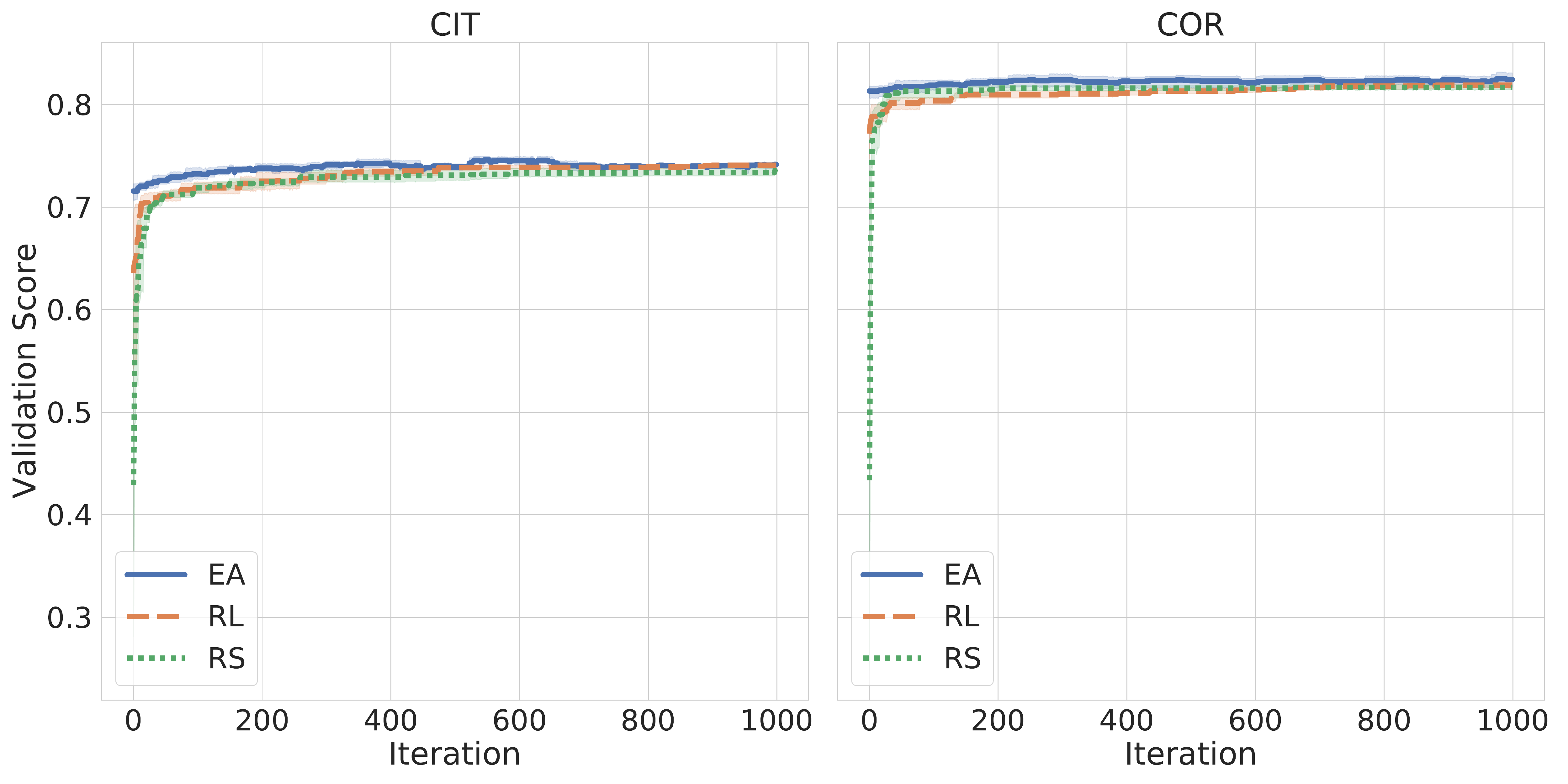}
    \caption{Highest validation accuracy by iteration, for \textbf{CIT} and \textbf{COR} datasets, on the \textbf{Macro} search space.}
    \label{fig:EV_RL_RS_highest}
\end{figure}

Table~\ref{tab:acc_time_table} shows the results of accuracy and execution time for the \matheus{Macro} and \matheus{Micro} search spaces, \matheus{at the end of the optimization process (after 1000 iterations)}. In terms of accuracy, the results obtained by the EA and RL methods are very similar to the ones obtained by the random search. \matheus{In terms of execution time, RS wins in most cases. The execution time for the search varies between 2 and 12 GPU hours.}

\begin{figure}[ht]
\center
    \includegraphics[width=\textwidth]{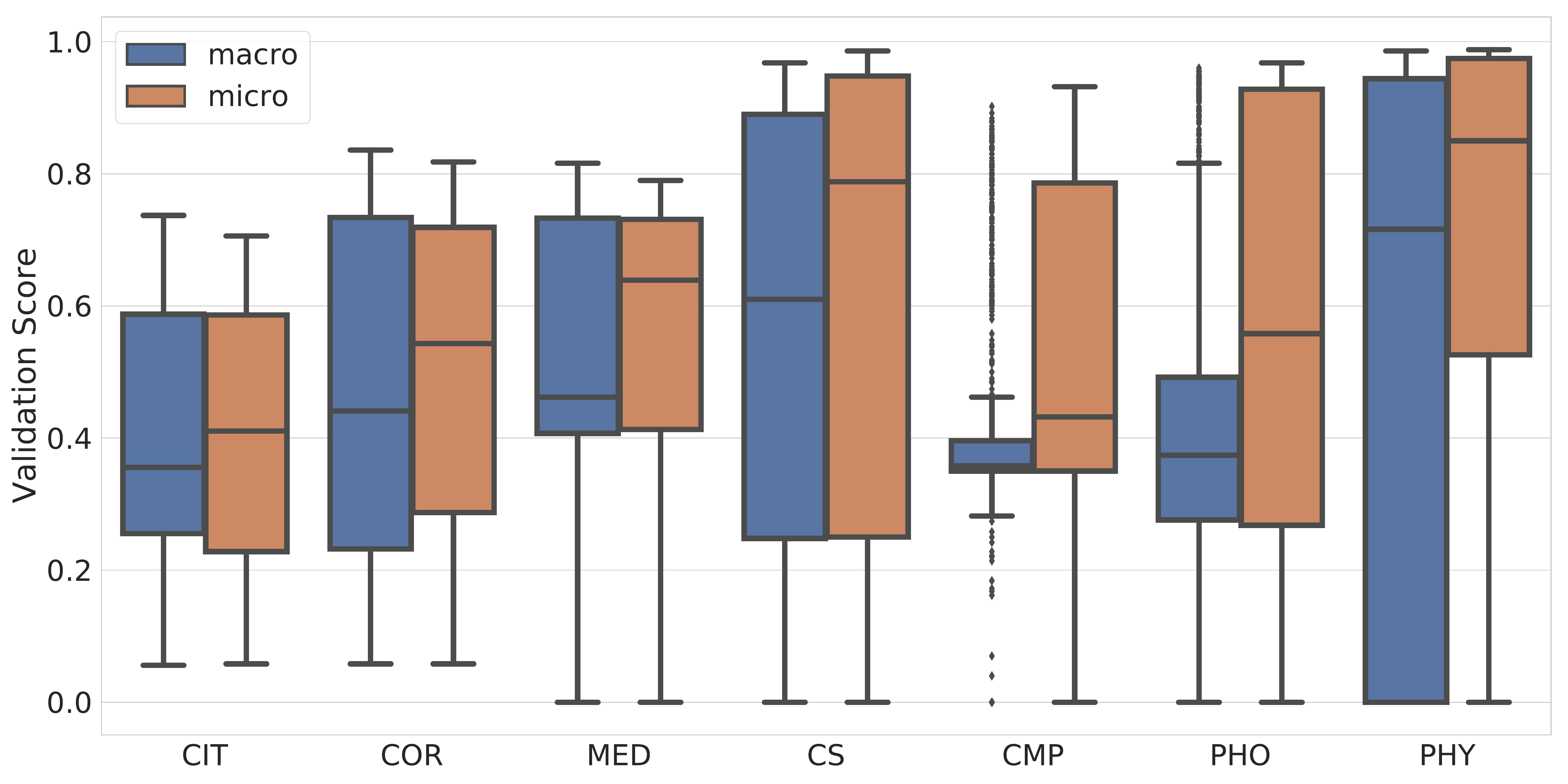}
    \caption{Distribution of EA's initial population validation accuracies on both search spaces.}
    \label{fig:initial_population}
\end{figure}

\matheus{Figure \ref{fig:EV_RL_RS_highest} presents the evolution of the highest validation accuracy value achieved by an GNN architecture across the iterations, by search method\footnote{We present only the results for the \textbf{Macro} search space because the results for \textbf{Micro} are very similar.}. Each line represents the mean validation score across all seeds, and the shaded area around it represents the standard deviation of this value. It is very clear that \textbf{all methods converge} (find a good performing architecture and plateaus) \textbf{within only a few iterations}. The fact that the EA already starts at a high value may be attributed to the population initialization process, depicted in Figure \ref{fig:initial_population}.}

It may seem counter-intuitive that we are using sophisticated methods to obtain results that can be also be achieved by a random search method, but as the authors in \cite{bergstra2012random} have previously discussed, in large search spaces where many of the dimensions are irrelevant to the task at hand the random search can be as effective as more sophisticated methods. This problem is aggravated by the neutrality of the space, i.e., architectures in neighbour regions of the search space may differ in a few components but do not lead to a value of accuracy different from their neighbors \cite{pimenta2020}. \matheus{Another stronger indicator of a neutral search space is the fact that many high quality individuals are generated in the initialization step, and evolution takes a minor part in improving them, as shown in Figure \ref{fig:EV_RL_RS_highest}.}

\begin{figure}[ht]
    \centering
    \includegraphics[width=\textwidth]{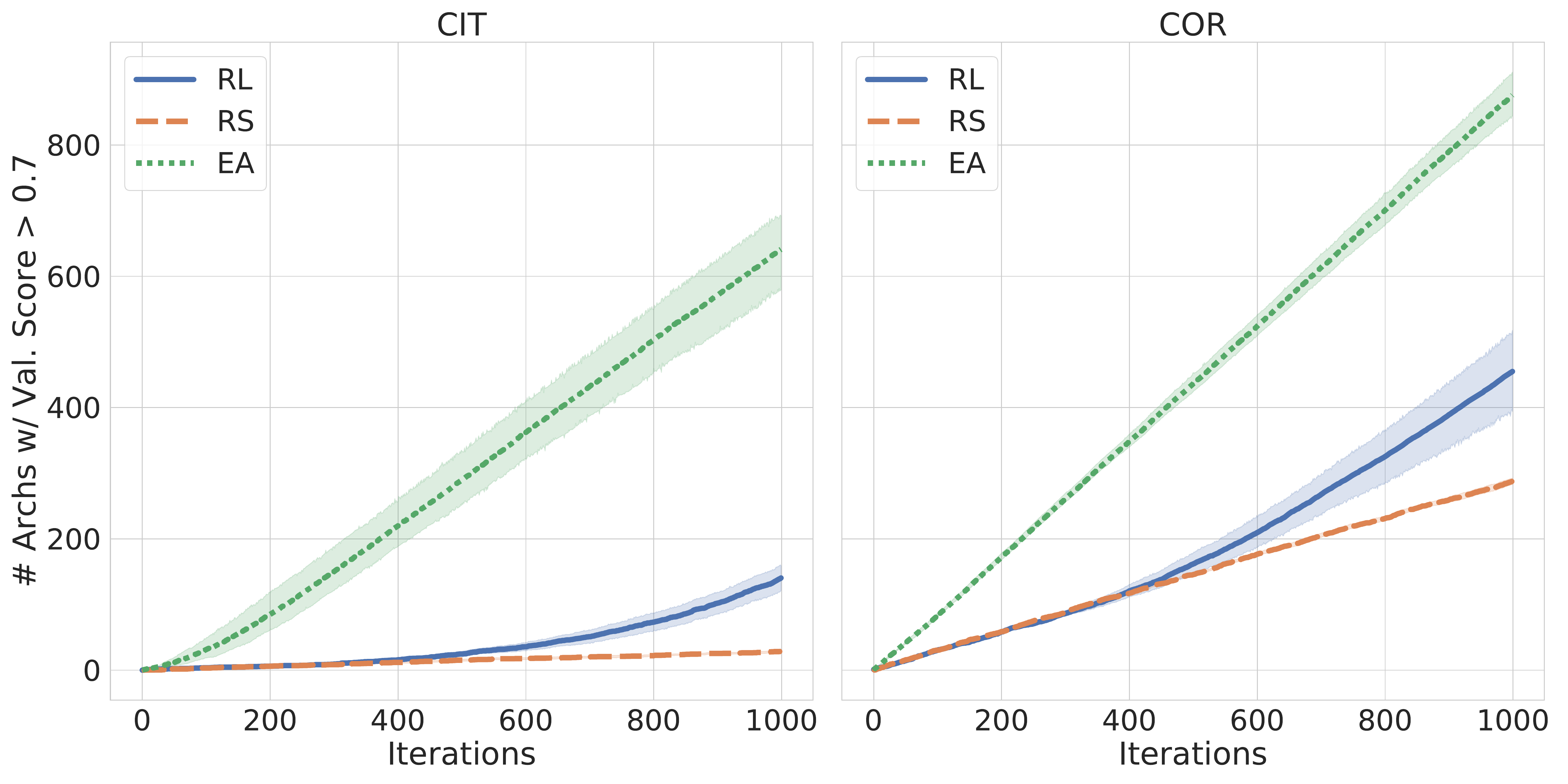}
    \caption{Cumulative number of architectures with validation accuracy higher than threshold, for \textbf{CIT} and \textbf{COR} datasets, on the \textbf{Macro} search space.}
    \label{fig:EV_RL_RS}
\end{figure}

Figure \ref{fig:EV_RL_RS} presents the \matheus{number of evaluated architectures with validation accuracy over $0.7$,}
for \textbf{CIT} and \textbf{COR}, in the Macro search space. \matheus{The  $0.7$ threshold was set because this value represents approximately the best accuracy value for \textbf{CIT} on the \textbf{Macro} search space. The pattern shown in the figure is consistent for \textbf{all datasets} in \textbf{both search spaces}.}
\matheus{
It shows that the EA tends to converge to a better region of the search space faster than the other two methods, thus evaluating more high quality architectures. Such tendency could be explained by the EA's selective pressure (driven by the tournament selection process), which makes the algorithm prioritize good individuals for mutation and evaluation.
}

\matheus{
The parameter size of GNNs is dependent on the dataset (since the structure of the neural network follows the graph) and on the choice of architecture.
Table \ref{tab:perct_GPU_oom} presents the percentage of generated architectures which exceeded GPU memory, by each dataset and search method.\footnote{The smallest datasets (\textbf{CIT} and \textbf{COR}) are not present in the table because none of the generated architectures for these datasets exceeded GPU memory.
}
EA is consistently the search method for which the \textbf{smallest percentage} of generated architectures are too big for the GPU memory, with the highest value as $16\%$, while RL reaches $81\%$ of all architectures being too large. This corroborates the findings of Real et. al. \cite{real2019aging} which state that Evolutionary Algorithms are able to find less complex but equally well performing architectures than RL.
}

\begin{table}[ht]
    \centering
    \caption{Percentages of generated architectures which exceeded the GPU memory and therefore were not evaluated, by dataset and search method}
    \label{tab:perct_GPU_oom}
    \begin{tabular}{llrr}
    \toprule
                 &             &  \textbf{Avg.}    &  \textbf{Max}  \\
                 &             &  \textbf{\%}      &  \textbf{\%}   \\
    \midrule
                 & \textbf{EA} &   $0.60 \pm  0.89$ &   $2.0$ \\
    \textbf{MED} & \textbf{RL} &   $3.20 \pm  0.84$ &   $4.0$ \\
        		 & \textbf{RS} &   $2.80 \pm  0.84$ &   $4.0$ \\\hline
        		 & \textbf{EA} &   $4.60 \pm  1.52$ &   $6.0$ \\
    \textbf{CS}  & \textbf{RL} &  $10.20 \pm  2.59$ &  $14.0$ \\
        		 & \textbf{RS} &   $9.60 \pm  1.52$ &  $11.0$ \\\hline
        		 & \textbf{EA} &  $13.60 \pm  1.82$ &  $\textbf{16.0}$ \\
    \textbf{PHY} & \textbf{RL} &  $41.80 \pm  9.44$ &  $56.0$ \\
        		 & \textbf{RS} &  $47.80 \pm  0.45$ &  $48.0$ \\\hline
        		 & \textbf{EA} &  $11.60 \pm  2.61$ &  $14.0$ \\
    \textbf{CMP} & \textbf{RL} &  $47.00 \pm 20.94$ &  $\textbf{81.0}$ \\
        		 & \textbf{RS} &  $38.40 \pm  1.67$ &  $41.0$ \\\hline
        		 & \textbf{EA} &   $4.60 \pm  2.70$ &   $9.0$ \\
    \textbf{PHO} & \textbf{RL} &  $20.80 \pm  3.42$ &  $24.0$ \\
        		 & \textbf{RS} &  $11.80 \pm  1.48$ &  $14.0$ \\
    \bottomrule
    \end{tabular}
\end{table}

\section{Conclusions and Future Work} \label{sec:conclusion}

GNNs are able to achieve state-of-the-art performances in prediction tasks over networks. However, their design and optimization is currently hand-made and error prone.
This paper compared the results of two NAS search methods -- a reinforcement learning technique \matheus{and} an evolutionary algorithm -- \matheus{to} a random search \matheus{in the task of searching} for architectures and hyperparameters for GNNs. 

The three methods \matheus{produced GNN architectures which achieved} similar results \matheus{in terms} of accuracy when considering a set of 7 datasets \matheus{and two architecture layer search spaces}, with the random search being the fastest method followed by the evolutionary algorithm and reinforcement learning. \matheus{Architectures generated by EA tend to fit in GPU memory, while the other methods generate oversized architectures in up to $\textbf{80\%}$ of cases. This shows that EA generates less complex structures while achieving a similar accuracy value to the other methods, corroborating the findings of Real et. al. \cite{real2019aging} for images.}

\matheus{In general, the} results indicate that there are irrelevant dimensions to this task in the defined search spaces, which will require a more in-depth study of each of these spaces.
Further, the neutrality of this space, i.e., the fact that neighbor solutions present different architectures but very similar results of accuracy make search even harder.
As future work, we intend to perform a more in-depth investigation of the dimensions of the search space in order to identify those that may be irrelevant to search, as well as propose new search methods that may include mechanisms to try to avoid these neutral regions.



\bibliographystyle{splncs04}
\bibliography{srcs/bibliography}

\begin{thebibliography}{10}
\providecommand{\url}[1]{\texttt{#1}}
\providecommand{\urlprefix}{URL }
\providecommand{\doi}[1]{https://doi.org/#1}

\bibitem{badrinarayanan2017segnet}
Badrinarayanan, V., Kendall, A., Cipolla, R.: Segnet: A deep convolutional
  encoder-decoder architecture for image segmentation. TPAMI'17
  \textbf{39}(12),  2481--2495 (2017)

\bibitem{bergstra2012random}
Bergstra, J., Bengio, Y.: Random search for hyper-parameter optimization.
  JMLR'12  \textbf{13}(Feb),  281--305 (2012)

\bibitem{bruna2014spectral}
Bruna, J., Zaremba, W., Szlam, A., LeCun, Y.: Spectral networks and locally
  connected networks on graphs. In: Bengio, Y., LeCun, Y. (eds.) ICLR'14 (2014)

\bibitem{cai2018nas}
Cai, H., Chen, T., Zhang, W., Yu, Y., Wang, J.: Efficient architecture search
  by network transformation. In: AAAI'18 (2018)

\bibitem{deferrard2016localized}
Defferrard, M., Bresson, X., Vandergheynst, P.: Convolutional neural networks
  on graphs with fast localized spectral filtering. In: Lee, D.D., Sugiyama,
  M., von Luxburg, U., Guyon, I., Garnett, R. (eds.) NeurIPS'16. pp. 3837--3845
  (2016)

\bibitem{eiben2015}
Eiben, A., Smith, J.: Introduction to Evolutionary Computing. Springer (2015)

\bibitem{elshawi2019automated}
Elshawi, R., Maher, M., Sakr, S.: Automated machine learning: State-of-the-art
  and open challenges. arXiv preprint arXiv:1906.02287  (2019)

\bibitem{elsken2019neural}
Elsken, T., Metzen, J.H., Hutter, F.: Neural architecture search: {A} survey.
  JMLR'19  \textbf{20},  55:1--55:21 (2019)

\bibitem{gao2019graphnas}
Gao, Y., Yang, H., Zhang, P., Zhou, C., Hu, Y.: Graph neural architecture
  search. In: IJCAI'20. pp. 1403--1409 (2020)

\bibitem{gori2005graphlearning}
Gori, M., Monfardini, G., Scarselli, F.: A new model for learning in graph
  domains. In: Proceedings. 2005 IEEE International Joint Conference on Neural
  Networks, 2005. vol.~2, pp. 729--734. IEEE (2005)

\bibitem{hamilton2017inductive}
Hamilton, W., Ying, Z., Leskovec, J.: Inductive representation learning on
  large graphs. In: NIPS '17 (2017)

\bibitem{he2016deep}
He, K., Zhang, X., Ren, S., Sun, J.: Deep residual learning for image
  recognition. In: CVPR'16. pp. 770--778 (2016)

\bibitem{kipf2017gcn}
Kipf, T.N., Welling, M.: Semi-supervised classification with graph
  convolutional networks. In: ICLR'17 (2017)

\bibitem{liu2016ssd}
Liu, W., Anguelov, D., Erhan, D., Szegedy, C., Reed, S., Fu, C.Y., Berg, A.C.:
  Ssd: Single shot multibox detector. In: ECCV'16. pp. 21--37. Springer (2016)

\bibitem{pimenta2020}
Pimenta, C.G., de~S{\'a}, A.G., Ochoa, G., Pappa, G.L.: Fitness landscape
  analysis of automated machine learning search spaces. In: EvoCOP'20. pp.
  114--130. Springer (2020)

\bibitem{real2019aging}
Real, E., Aggarwal, A., Huang, Y., Le, Q.V.: Aging evolution for image
  classifier architecture search. In: AAAI'19 (2019)

\bibitem{scarselli2009neural}
Scarselli, F., Gori, M., Tsoi, A.C., Hagenbuchner, M., Monfardini, G.: The
  graph neural network model. {IEEE} TNN'09  (2009)

\bibitem{sen2008collective}
Sen, P., Namata, G., Bilgic, M., Getoor, L., Galligher, B., Eliassi-Rad, T.:
  Collective classification in network data. AI magazine  \textbf{29}(3),
  93--93 (2008)

\bibitem{shchur2018pitfalls}
Shchur, O., Mumme, M., Bojchevski, A., G{\"u}nnemann, S.: Pitfalls of graph
  neural network evaluation. arXiv preprint arXiv:1811.05868  (2018)

\bibitem{velickovic2018gat}
Velickovic, P., Cucurull, G., Casanova, A., Romero, A., Li{\`{o}}, P., Bengio,
  Y.: Graph attention networks. In: ICLR'18 (2018)

\bibitem{wu2019comprehensive}
Wu, Z., Pan, S., Chen, F., Long, G., Zhang, C., Yu, P.S.: A comprehensive
  survey on graph neural networks. CoRR  (2019)

\bibitem{xu2018powerful}
Xu, K., Hu, W., Leskovec, J., Jegelka, S.: How powerful are graph neural
  networks? In: ICLR'19. OpenReview.net (2019)

\bibitem{zhang2018end}
Zhang, M., Cui, Z., Neumann, M., Chen, Y.: An end-to-end deep learning
  architecture for graph classification. In: AAAI'18 (2018)

\bibitem{zhang2020gnn}
{Zhang}, Z., {Cui}, P., {Zhu}, W.: Deep learning on graphs: A survey. TKDE'20
  pp.~1--1 (2020)

\bibitem{zhou2019autognn}
Zhou, K., Song, Q., Huang, X., Hu, X.: Auto-gnn: Neural architecture search of
  graph neural networks. arXiv preprint arXiv:1909.03184  (2019)

\bibitem{zoph2018image}
Zoph, B., Vasudevan, V., Shlens, J., Le, Q.V.: Learning transferable
  architectures for scalable image recognition. In: CVPR'2018 (2018)

\end{thebibliography}

\end{document}